\documentclass{article}

\usepackage[preprint, nonatbib]{neurips_2023}
\usepackage[numbers,sort]{natbib} 

\usepackage[utf8]{inputenc} 
\usepackage[T1]{fontenc}    
\usepackage{hyperref}       
\usepackage{url}            
\usepackage{booktabs}       
\usepackage{amsfonts}       
\usepackage{nicefrac}       
\usepackage{microtype}      
\usepackage{xcolor}         

\usepackage{algorithm,algpseudocode}
\usepackage{algorithmicx}

\usepackage{graphicx}
\usepackage{amsmath}
\usepackage{amssymb}

\usepackage{multirow}
\usepackage{multicol}
\usepackage{lineno}
\usepackage{booktabs}
\usepackage{caption}

\usepackage{graphicx}
\usepackage{amsmath}
\usepackage{amssymb}
\usepackage{booktabs}

\usepackage{sidecap}

\usepackage{algorithm,algpseudocode}
\usepackage{algorithmicx}

\usepackage[normalem]{ulem}
\usepackage{ifthen}

\usepackage{xcolor}
\usepackage{caption}
\usepackage{subcaption}

\newcommand*\samethanks[1][\value{footnote}]{\footnotemark[#1]}
\newcommand{\modelname}{\emph{Photoswap}}
\newcommand{\bs}[1]{{\boldsymbol{#1}}}
\newcommand{\getNoise}{\bs{\epsilon}_{\theta^*}}

\usepackage[capitalize]{cleveref}
\crefname{section}{Sec.}{Secs.}
\Crefname{section}{Section}{Sections}
\Crefname{table}{Table}{Tables}
\crefname{table}{Tab.}{Tabs.}

\definecolor{amber}{rgb}{1.0, 0.4, 0.0}

\title{\textsc{Photoswap}: \\ Personalized Subject Swapping in Images}

\author{
Jing Gu$^{1}$\thanks{~Correspondence to Jing Gu and Xin Eric Wang, \texttt{\{jgu110,xwang366\}@ucsc.edu}.} \quad Yilin Wang$^{2}$ \quad Nanxuan Zhao$^{2}$ \quad Tsu-Jui Fu$^{3}$ \quad Wei Xiong$^{2}$ \quad Qing Liu$^{2}$ \\
\textbf{Zhifei Zhang$^{2}$ \quad He Zhang$^{2}$ \quad Jianming Zhang$^{2}$ \quad  HyunJoon Jung$^{2}$ \quad Xin Eric Wang$^{1}$\samethanks[1] } \\ 
\textsuperscript{1}University of California, Santa Cruz\; \\
\textsuperscript{2}Adobe \;
\textsuperscript{3}University of California, Santa Barbara
\and
\url{https://photoswap.github.io/}
}
\begin{document}

\maketitle

\begin{figure}[H]
\vspace{-4ex}
    \centering
    \includegraphics[width=\textwidth]{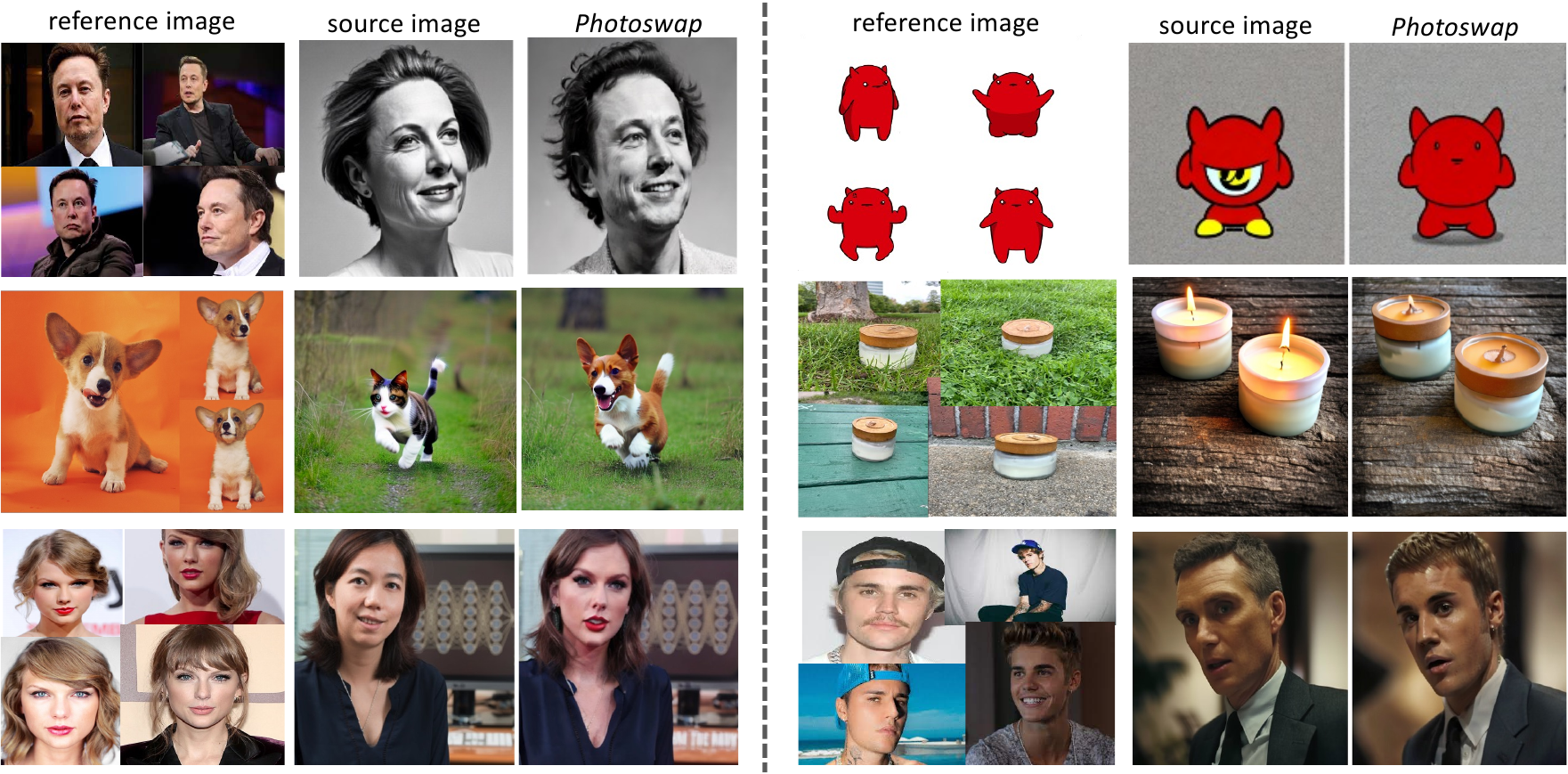}
        \caption{
        \modelname{} can effortlessly replace the subject in a source image, which could be either synthetic (first two rows) or real (bottom row), with a personalized subject specified in reference images, while preserving the original subject pose and the composition of the source image. 
        }
        \label{fig:real_img_swapping}
\end{figure}

\begin{abstract}
In an era where images and visual content dominate our digital landscape, the ability to manipulate and personalize these images has become a necessity.
Envision seamlessly substituting a tabby cat lounging on a sunlit window sill in a photograph with your own playful puppy, all while preserving the original charm and composition of the image. 
We present \emph{Photoswap}, a novel approach that enables this immersive image editing experience through personalized subject swapping in existing images.
\emph{Photoswap} first learns the visual concept of the subject from reference images and then swaps it into the target image using pre-trained diffusion models in a training-free manner. We establish that a well-conceptualized visual subject can be seamlessly transferred to any image with appropriate self-attention and cross-attention manipulation, maintaining the pose of the swapped subject and the overall coherence of the image. 
Comprehensive experiments underscore the efficacy and controllability of \emph{Photoswap} in personalized subject swapping. Furthermore, \emph{Photoswap} significantly outperforms baseline methods in human ratings across subject swapping, background preservation, and overall quality, revealing its vast application potential, from entertainment to professional editing.
\end{abstract}

\section{Introduction}
\label{sec:intro}

Imagine a digital world where the boundaries of reality and creativity blur, where a photograph of a tabby cat lounging in a sunlit window sill can effortlessly be transformed to feature your playful puppy in the same pose. Or envision yourself as a part of a famous movie scene, replaced seamlessly with the original character while preserving the very essence and composition of the scene. Can we achieve this level of personalized image editing, not just with expert-level photo manipulation skills, but in an automated, user-friendly manner? This question lies at the heart of \emph{personalized subject swapping}, the challenging task of replacing the subject in an image with a user-specified subject, while maintaining the integrity of the original pose and composition. It opens up a plethora of applications in areas such as entertainment, advertising, and professional editing.

Personalized subject swapping is a complex undertaking that comes with its own set of challenges. The task requires a profound comprehension of the visual concept inherent to both the original subject and the replacement subject. Simultaneously, it demands the seamless integration of the new subject into the existing image. One of the critical objectives in subject swapping is to preserve the similar pose of the replacement subject. It is crucial that the swapped subject seamlessly fits into the original pose and scene, creating a natural and harmonious visual composition. This necessitates careful consideration of factors such as lighting conditions, perspective, and overall aesthetic coherence. By effectively blending the replacement subject with these elements, the final image maintains a sense of continuity and authenticity.

Existing image editing methods fall short in addressing these challenges. Many of these techniques are restricted to global editing and lack the finesse needed to seamlessly integrate new subjects into existing images. For example, for most text-to-image (T2I) models, a slightly prompt change could lead to a totally different image. Recent works  \cite{nichol2021glide,meng2021sdedit,couairon2022diffedit,cao2023masactrl,zhang2023adding} allow user to control the generation with an additional input such as user brush, semantic layout, or sketches. However, it is still challenging to guide the generation process to follow users' intent on the generation of object shape, texture, and identity. Other approaches~\cite{hertz2022prompt,tumanyan2022plug,mokady2022null-text} have explored the potential of using text prompts to edit image content in the context of synthetic image generation. Despite showing promise, these methods are not yet fully equipped to handle the intricate task of swapping subjects in existing images with user-specified subjects.

Therefore, we present \emph{Photoswap}, a novel framework that leverages pre-trained diffusion models for personalized subject swapping in images. In our approach, the diffusion model learns to represent the concept of the subject ($O_t$). Then the representative attention map and attention output saved in the source image generation process will be transferred into the generation process of the target image to generate the new subject while keeping non-subject pixels unchanged.
Our extensive experiments and evaluations demonstrate the effectiveness of \emph{Photoswap}. Not only does our method enable the seamless swapping of subjects in images, but it also maintains the pose of the swapped subject and the overall coherence of the image. Remarkably, \emph{Photoswap} outperforms baseline methods by a large margin in human evaluations of subject identity preservation, background preservation, and overall quality of the swapping (\textit{e.g.}, 50.8\% \textit{vs.} 28.0\% in terms of overall quality). The contributions of this work are as follows: 
\textbf{1)} We present a new framework for personalized subject swapping in images.
\textbf{2)} We propose a training-free attention swapping method that governs the editing process.
\textbf{3)} The efficacy of our proposed framework is demonstrated through extensive experiments including human evaluation.

\section{Related Work}
\label{sec:related-work}

\subsection{Text-to-Image Generation}

In the early stages of text-based image generation, Generative Adversarial Networks (GANs)~\cite{goodfellow2020generative,brock2018large,karras2019style} were widely used due to their exceptional ability to produce high-quality images. These models aimed to align textual descriptions with synthesized images through multi-modal vision-language learning, achieving impressive results on specific domains (e.g., bird, chair and human face). When combined with CLIP \cite{radford2021learning}, a large pre-trained model that learns visual-textual representations from millions of caption-image pairs, GAN models \cite{crowson2022vqgan} have demonstrated promising outcomes in cross-domain text-to-image (T2I) generation.
Recently, T2I generation has seen remarkable progress with auto-regressive~\cite{dalle, ding2021cogview, ding2021cogview} and diffusion models~\cite{nichol2021glide, gu2022vector, dalle2, saharia2022photorealistic}, offering diverse outcomes and can synthesize high-quality images closely aligned with textual descriptions in arbitrary domains.


Rather than focusing on T2I generation tasks without any constraints, subject-driven T2I generation~\cite{nitzan2022my-style, casanova2021ic-gan, ruiz2023dream-booth} requires the model to identify the specific object from a set of visual examples and synthesize novel scenes incorporating them based on the input text prompts. Building upon modern diffusion techniques, recent approaches such as DreamBooth~\cite{ruiz2023dream-booth} and Textual Inversion~\cite{gal2023ti,gal2023edt,kumari2023mcc,mokady2022null-text} learn to invert special tokens from a given set of images. By combining these tokens with text prompts, they generate personalized unseen images. To improve data efficiency, retrieval augmentation techniques~\cite{sheynin2023knn-diffusion,blattmann2022retrieval-diffusion,chen2023re-imagen} leverages external knowledge bases to overcome limitations posed by rare entities, resulting in visually relevant appearances and enhanced personalization.
In our work, we aim to tackle personalized subject swapping, not only preserving the identity of subjects in reference images, but also maintaining the context of the source image.

\subsection{Text-guided Image Editing}

Text-guided image editing manipulates an existing image based on the input textual instructions, while preserving certain aspects or characteristics of the original image. Early works based on GAN models~\cite{karras2019style} only limits to a certain object domain. Diffusion-based methods~\cite{zhang2023adding,nichol2021glide,feng2023training-free} break this barrier and support text-guided image editing. Though these methods generate stunning results, many of them suffer from conducting local editing, and additional manual masks~\cite{meng2021sdedit,zeng2022scenecomposer,meng2021sdedit} are required to constrain the editing regions, which is often tedious to draw. By employing cross-attention \cite{hertz2022prompt} or spatial characteristics \cite{tumanyan2022plug}, the local editing can be achieved but struggles with non-rigid transformations (e.g., changing pose) and retaining the original image layout structure. While Imagic~\cite{kawar2022imagic} addresses the need for non-rigid transformations by fine-tuning a pre-trained diffusion model to capture image-specific appearances, it requires test-time finetuning, which is not time-efficient for deployment. Moreover, relying solely on text as input lacks precise control. In contrast, we propose a novel training-free attention swapping scheme that enables precise personalization based on reference images, without the need for time-consuming finetuning.

\subsection{Exemplar-guided Image Editing}

Exemplar-guided image editing covers a broad range of applications, and most of the works~\cite{wang2019example,huang2018multimodal,zhou2021cocosnet} can be categorized as exemplar-based image translation tasks, conditioning on various information, such as stylized images~\cite{liu2021adaattn, deng2022stytr2, zhang2022inversion}, layouts~\cite{yang2022reco, li2023gligen, jahn2021high}, skeletons~\cite{li2023gligen}, sketches/edges~\cite{seo2022midms}. With the convenience of stylized images, image style transfer~\cite{liao2017visual,zhang2020cross} receives extensive attentions, replying on methods to build a dense correspondence between input and reference images, but it cannot deal with local editing. To achieve local editing with non-rigid transformation, conditions like bounding boxes and skeletons are introduced, but require drawing efforts from users, which sometimes are hard to obtain. A recent work~\cite{yang2022paint} poses exemplar-guided image editing task as an inpainting task with the mask and transfers the semantic content from the reference image to the source one, with the context intact. Different from these works, we propose a more user-friendly scenario by conducting personalized subject swapping with only reference images and obtain high-quality editing results.

\section{Preliminary}
\label{sec:preliminary}

Diffusion models are a type of generative model that operates probabilistically. In this process, an image is created by gradually eliminating noise from the target that is characterized by Gaussian noise.
In the context of text-to-image generation, a diffusion model typically involves a process where an initial random image is gradually refined step by step, with each step guided by a learned model, until it becomes a realistic image. The changes to the image spread out and affect many pixels over time. Given an initial random noise $\bs{z}_T\!\sim\!\mathcal{N}(0, \mathbf{I})$, the diffusion model gradually denoise $\bs{z}_t$, which gives $\bs{z}_{t-1}$. 

Diffusion models are probabilistic generative models that learn to generate images by simulating a random process called a diffusion process.
In the image generation process, the diffusion model gradually predicts the noise at the current diffusion step and denoises to get the final image. 
In this study, we utilize a pre-trained text-to-image diffusion model, Stable Diffusion~\cite{rombach2022high}, which encodes the image into latent space and gradually denoises the latent variable to generate a new image. Stable Diffusion is based on a U-Net architecture~\cite{ronneberger2015u}, which generates latent variable $\bs{z}_{t-1}$ conditioned on a given text prompt $P$ and the latent variable $\bs{z}_t$ from the previous step $t$:
\begin{equation}
    \bs{z}_{t-1} = \bs{\epsilon_\theta}(\bs{z}_t, P, t)
    \label{eq:injection}
\end{equation}
 The U-Net consists of layers that include repetition of self-attention and cross-attention blocks. This study focuses on manipulating self-attention and cross-attention to achieve the task of personalized subject swapping.






\section{The \emph{Photoswap} Method}
\label{sec:method}

\begin{figure*}[t!]
    \centering
    \includegraphics[width=\linewidth]{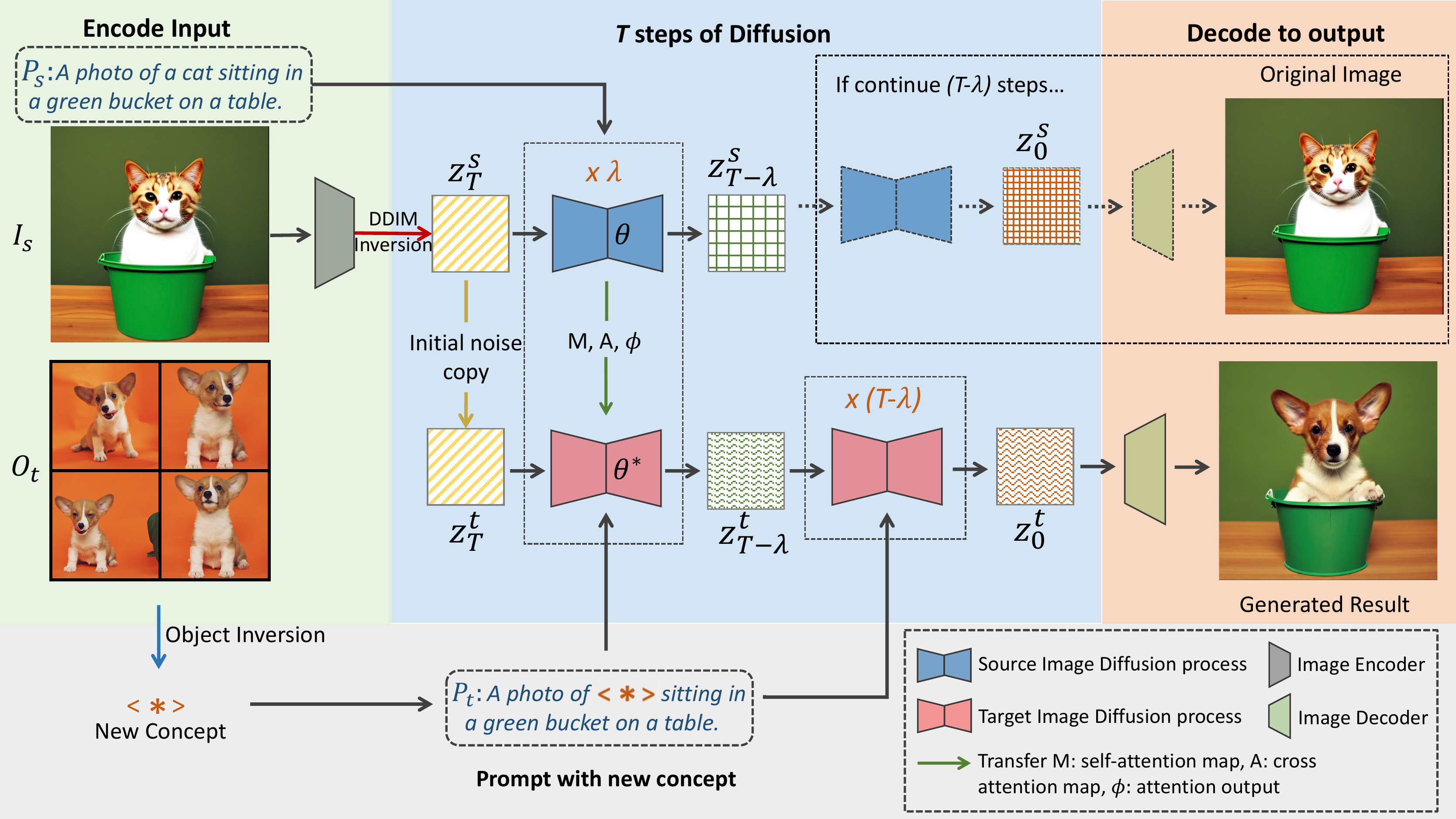}
    \caption{\textbf{The \emph{Photoswap} framework.} Given several images of a new concept, the diffusion model first learns the concept and converts it into a token. The upper part is the generation process of the source image, while the bottom part is the generation process of target image. The initial noise feature $z^{t}_{T}$ is copied from $z^{s}_{T}$ of the source. 
    The attention output and attention map in the source image generation process would be transferred to the target image generation process. The final feature $z^{t}_{0}$ is decoded to output the target image.
    Refer to Section~\ref{sec:method} for more details.
    }
    \label{fig:architecture}
\end{figure*}

Providing a few reference images of a personalized target subject $O_t$, \emph{Photoswap} can seamlessly swap it with another subject $O_s$ in a given source image $I_s$. The \emph{Photoswap} pipeline is illustrated in Figure~\ref{fig:architecture}. To learn the visual concept of the target subject $O_t$, we fine-tune a diffusion model with reference images and do object inversion to represent $O_t$ using special token \textbf{*}.
Then, to substitute the subject in the source image, we first obtain the noise $z_T$ \footnote{For a synthetic image, $z_T^*$ is the initial noise used to generate it. For a real image, we utilize an improved version of DDIM inversion~\cite{song2020denoising} to get the initial noise and re-generate the source image. See Sec.~\ref{sec:implementation} for details.} that can be used to re-construct the source image $I_s$. 
Next, through the U-Net, we obtain the needed feature map and attention output in the self-attention and cross-attention layers, including $M$, $A$, and $\bs{\phi}$ (which we will introduce in Sec.~\ref{subsec:swapping}). 
Finally, during the target image generation process that is conditioned on the noise $z_T$ and the target text prompt $P_t$, in the first $\lambda$ steps, those intermediate variables ($M$, $A$, and $\bs{\phi}$) would be replaced with corresponding ones obtained during the the source image generation process. In the last ($T-\lambda$) steps, no attention swapping is needed and we can continue the denoising process as usual to obtain the final resulting image. Sec.~\ref{subsec:visual_concept_learning} discusses the visual concept learning technique we used, and Sec.~\ref{subsec:swapping} details the training-free attention swapping method for controllable subject swapping.

\subsection{Visual Concept Learning}
\label{subsec:visual_concept_learning}

Subject swapping requires a thorough understanding of the subject's identity and specific characteristics. This knowledge enables the creation of accurate representations that align with the source subject. The subject's identity influences the composition and perspective of the image, including its shape, proportions, and textures, which affect the overall arrangement of elements.
However, existing diffusion models lack information about the target subject ($O_t$) in their weights because the training data for text-to-image generation models does not include personalized subjects. To overcome this limitation and generate visually consistent variations of subjects from a given reference set, we need to personalize text-to-image diffusion models accurately. Recent advancements have introduced various methods, such as fine-tuning the diffusion model with distinct tokens associated with specific subjects, to achieve this ``personalization''\cite{gal2022image, ruiz2023dream-booth, kumari2023mcc}. In our experiments, we primarily utilize DreamBooth\cite{ruiz2023dream-booth} as a visual concept learning method. It's worth noting that alternative concept learning methods can also be effectively employed with our framework.

\begin{figure*}[t!]
    \centering
    \includegraphics[scale=0.7]{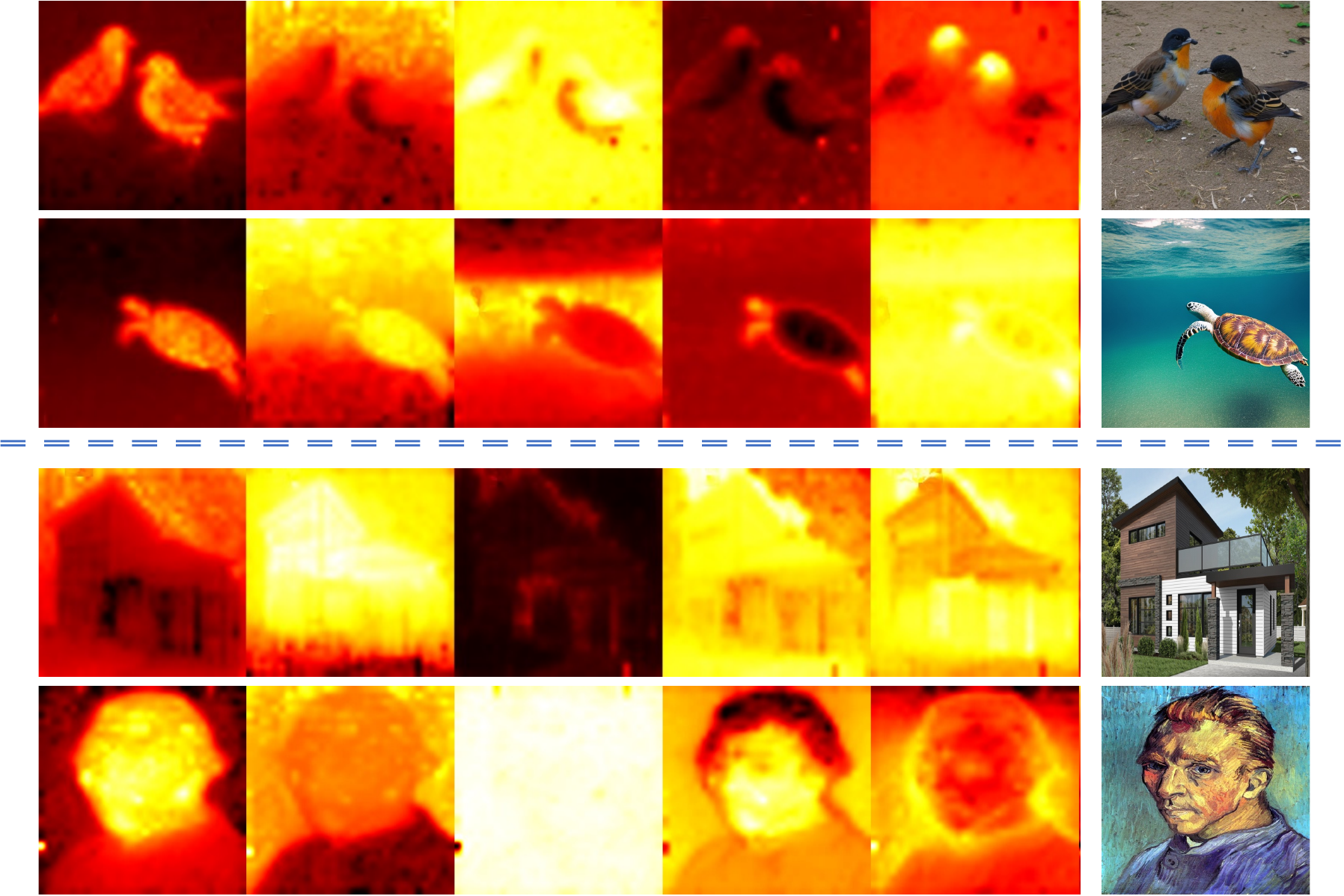}
    \caption{\textbf{SVD visualization of self-attention maps.} Each image's attention map is resized to 64x64 at every layer, and we calculate the average map across all layers for all diffusion time steps. Most significant components are extracted with SVD and visualized. Remarkably, the visualized results demonstrate a strong correlation with the layout of the generated image. The top two rows are visualization about synthetic images while the bottom two rows are about real images.
    }
    \label{fig:self-attention-map-visualization}
\end{figure*}

\begin{figure}
    \centering
    \includegraphics[scale=0.7]{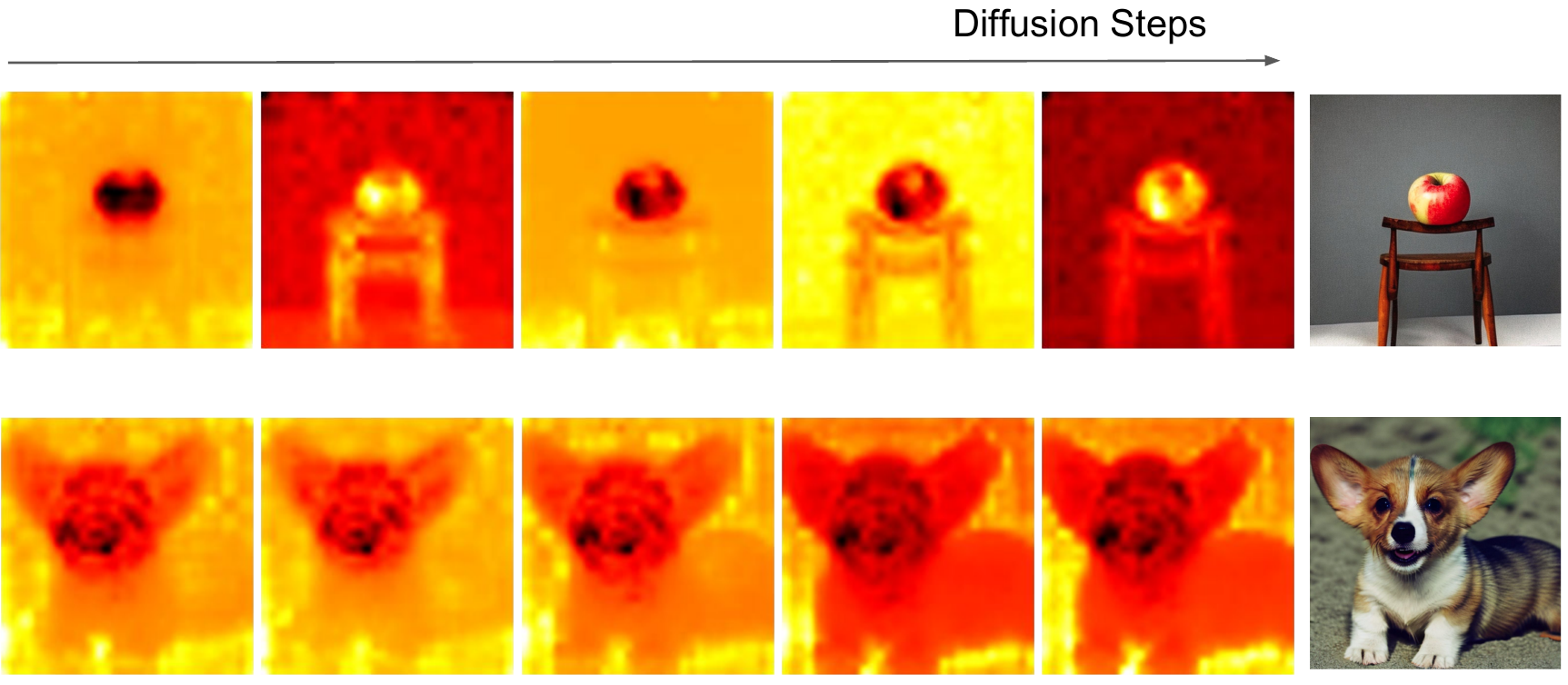}
    \caption{\textbf{Self-attention map visualization across diffusion time steps.} 
    This representation reveals that the layout of the generated image is intrinsically embedded in the self-attention map from the initial steps. Consequently, to assert control over the layout, it is imperative to commence the attention swap at the earliest stages of the process.
    }
    \label{fig:diffusion_step_map}
\end{figure}

\subsection{Controllable Subject Swapping via Training-free Attention Swapping}
\label{subsec:swapping}

Subject swapping poses intriguing challenges, requiring the maintenance of the source image's spatial layout and geometry while integrating a new subject concept within the same pose. This necessitates preserving the critical features in the source latent variable, which encapsulates the source image information, and leveraging the influence of the target image text prompt $P_t$, which carries the concept token, to inject the new subject into the image.

The central role of the attention layer in orchestrating the generated image's layout has been well-established in prior works~\cite{hertz2022prompt, cao2023masactrl, tumanyan2022plug}. To keep non-subject pixels intact, we orchestrate the generation of the target image $I_t$ by transferring vital variables to the target image generation process. Here, we explore how distinct intermediate variables within the attention layer can contribute to a controllable generation in the context of subject swapping.

Within the source image generation process, we denote the cross-attention map as $\bs{A}^s_i$, the self-attention map as $\bs{M}^s_i$, the cross-attention output as $\psi^s_i$, and the self-attention output as $\phi^s_i$. The corresponding variables in the target image generation process are denoted as $\bs{A}^t_i$, $\bs{M}^t_i$, $\psi^t_i$, $\phi^t_i$, where $i$ represents the current diffusion step.

In the self-attention block, the latent feature $z_i$ is projected into queries $\bs{q_i}$, keys $\bs{k_i}$, and values $\bs{v_i}$. We obtain the self-attention block's output $\bs{\phi}_i$ using the following equation:
\begin{equation}
\bs{\phi}_i = \bs{M}_i\bs{v}_i \quad \text{where} \quad \bs{M}_i = \text{Softmax}\left(\bs{q}_i {\bs{k}_i}^T \right)
\label{eq:selfattn}
\end{equation}
where $\bs{M}_i$ is the self-attention map, and $\phi_i$ is the feature output from the self-attention layer. 
The cross-attention block's output $\bs{\psi}_i$ is:
\begin{equation}
\bs{\psi}_i = \bs{A}_i\bs{v}_i \quad \text{where} \quad \bs{A}_i = \text{Softmax}\left(\bs{q}_i {\bs{k}_i}^T \right)
\label{eq:crossattention}
\end{equation}
where $\boldsymbol{A}_i$ is the cross-attention map. In both self-attention and cross-attention, the attention map $\boldsymbol{M}_i$ and $\boldsymbol{A}_i$ are correlated to the similarity between $q_i$ and $k_i$, acting as weights that dictate the combination of information in $v_i$.
In this work, the manipulation of the diffusion model focus on self-attention and cross-attention within U-Net, specifically, swapping $\boldsymbol{\phi}$, $\boldsymbol{M}$, and $\boldsymbol{A}$, while keeping $\psi$ unchanged.

{\small \begin{algorithm}[t!]
\caption{The \emph{Photoswap} Algorithm}\label{alg:pnp}
\begin{algorithmic}
\State \textbf{Inputs:} source image $I_s$, reference images $O_t$, source image text prompt $P_s$, target image text prompt $P_t$, diffusion model $\theta$ \\
\State $\theta^* \gets \theta, O_t$  $\triangleright$ Finetune diffusion model to include the new concept
\State $\boldsymbol{z}_T^s \gets DDIMInversion(ImageEncoder(I_s), P_s)$$\triangleright$ Using DDIM to guarantee re-construction
\State $\bs{z}^t_T \gets \boldsymbol{z}_T^s$ $\triangleright$ Using the same starting noise
\For{$i = T, T-1, ..., 1$}
  \State $\epsilon^s,\boldsymbol{\phi}^s_i, \boldsymbol{M}^s_i, \boldsymbol{A}^s_i \gets \getNoise\!(\boldsymbol{z}^s_i,P_s,i)$ $\triangleright$ Denoise to get the attention output and map for source image
  \State $\boldsymbol{\phi}_i^t, \boldsymbol{M}_i^t, \boldsymbol{A}^t_i \gets \getNoise\!\!\left(\boldsymbol{z}_i^t,P_t,i\right)$ $\triangleright$ Denoise to get the attention output and map for target image
  \State $\boldsymbol{\phi}_i^*, \boldsymbol{M}_i^*, \boldsymbol{A}^*_i \gets \text{SWAP}(\boldsymbol{\phi}_i^s, \boldsymbol{M}_i^s, \boldsymbol{A}^s_i, \boldsymbol{\phi}_i^t, \boldsymbol{M}_i^t, \boldsymbol{A}^t_i, i)$
  \State $\epsilon^* \gets \getNoise\!\!\left(\boldsymbol{z}_i^t,P_t,i, \boldsymbol{\phi}_i^*, \boldsymbol{M}_i^*, \boldsymbol{A}^*_i\right)$ $\triangleright$ Denoise the updated attention map and output
  \State $\boldsymbol{z}_{i-1}^s \gets DDIMSampler(\boldsymbol{z}_{i}^s, \epsilon^s)$ $\triangleright$ Sample next latent variable for source image
  \State $\boldsymbol{z}_{i-1}^t \gets DDIMSampler(\boldsymbol{z}_{i}^t, \epsilon^*)$ $\triangleright$ Sample next latent variable for source image
\EndFor
\State $I_t = ImageDecoder(\boldsymbol{z}_{0}^t)$  \\
\Return $I_t$
\\
\Function{SWAP}{$\boldsymbol{\phi}^s, \boldsymbol{M}^s, \boldsymbol{A}^s, \boldsymbol{\phi}^t, \boldsymbol{M}^t, \boldsymbol{A}^t, i$}
\State $\boldsymbol{\phi}^* \gets (i<\lambda_\phi) ? \boldsymbol{\phi}^s : \boldsymbol{\phi}^t$ $\triangleright$ Control self-attention feature swap
\State $\boldsymbol{M}^* \gets (i<\lambda_M) ? \boldsymbol{M}^s : \boldsymbol{M}^t$ $\triangleright$ Control self-attention Map swap
\State $\boldsymbol{A}^* \gets (i<\lambda_A) ? \boldsymbol{A}^s : \boldsymbol{A}^t$ $\triangleright$ Control cross-attention map swap \\
\Return$\boldsymbol{\phi}^*, \boldsymbol{M}^*, \boldsymbol{A}^*$
\EndFunction
\end{algorithmic}
\label{alg1}
\end{algorithm}}

\textbf{Self-attention map $\boldsymbol{M}$}, as it calculates the similarity within spatial features after linear projection, plays a pivotal role in governing spatial content during the generation process. As visualized in Figure~\ref{fig:self-attention-map-visualization}, we capture $\boldsymbol{M}$ during the image generation and highlight the leading components via Singular Value Decomposition (SVD). This visualization reveals a high correlation between $\boldsymbol{M}$ and the geometry and content of the generated image. Further, when visualizing the full steps of the diffusion process (Figure~\ref{fig:diffusion_step_map}), we discern that the layout information is mirrored in the self-attention from the initial steps. This insight underscores the necessity of initiating the swap early on to prevent the emergence of a new, inherent layout.

\textbf{Cross-attention map $\bs{A}$} is determined by both latent variable and text prompt, as in Equation~\ref{eq:crossattention}, and $\bs{A}^s_iv$ can be viewed as a weighted sum of the information from a text prompt. Copying $\bs{A}^s_i$ to $\bs{A}^t_i$ during the target image generation process improves the layout alignment between the source image and the target image.  

\textbf{Self-attention output $\phi$}, derived from the self-attention layer, encapsulates rich content information from the source image, independent of direct computation with textual features. Hence, replacing $\boldsymbol{\phi_i^t}$ with $\boldsymbol{\phi_i^s}$ enhances the preservation of context and composition from the original image. Our observations indicate that $\boldsymbol{\phi}$ exerts a more profound impact on the image layout than the cross-attention map $\boldsymbol{A}$.

\textbf{Cross-attention output $\psi$}, emanating from the cross-attention layer, embodies the visual concept of the target subject. It is vital to note that substituting cross-attention output $\psi_i^s$ with $\psi_i^t$ would obliterate all information from the target text prompt $P_t$, as illustrated in Equation~\ref{eq:crossattention}. Given that $k_i^t$ and $v_i^t$ are projections of target prompt embeddings, we retain $\psi_i^s$ unchanged to safeguard the target subject's identity.
 
Algorithm~\ref{alg1} provides the pseudo code of our full \emph{Photoswap} algorithm.

\begin{figure}
    \centering
    \includegraphics[width=\linewidth]{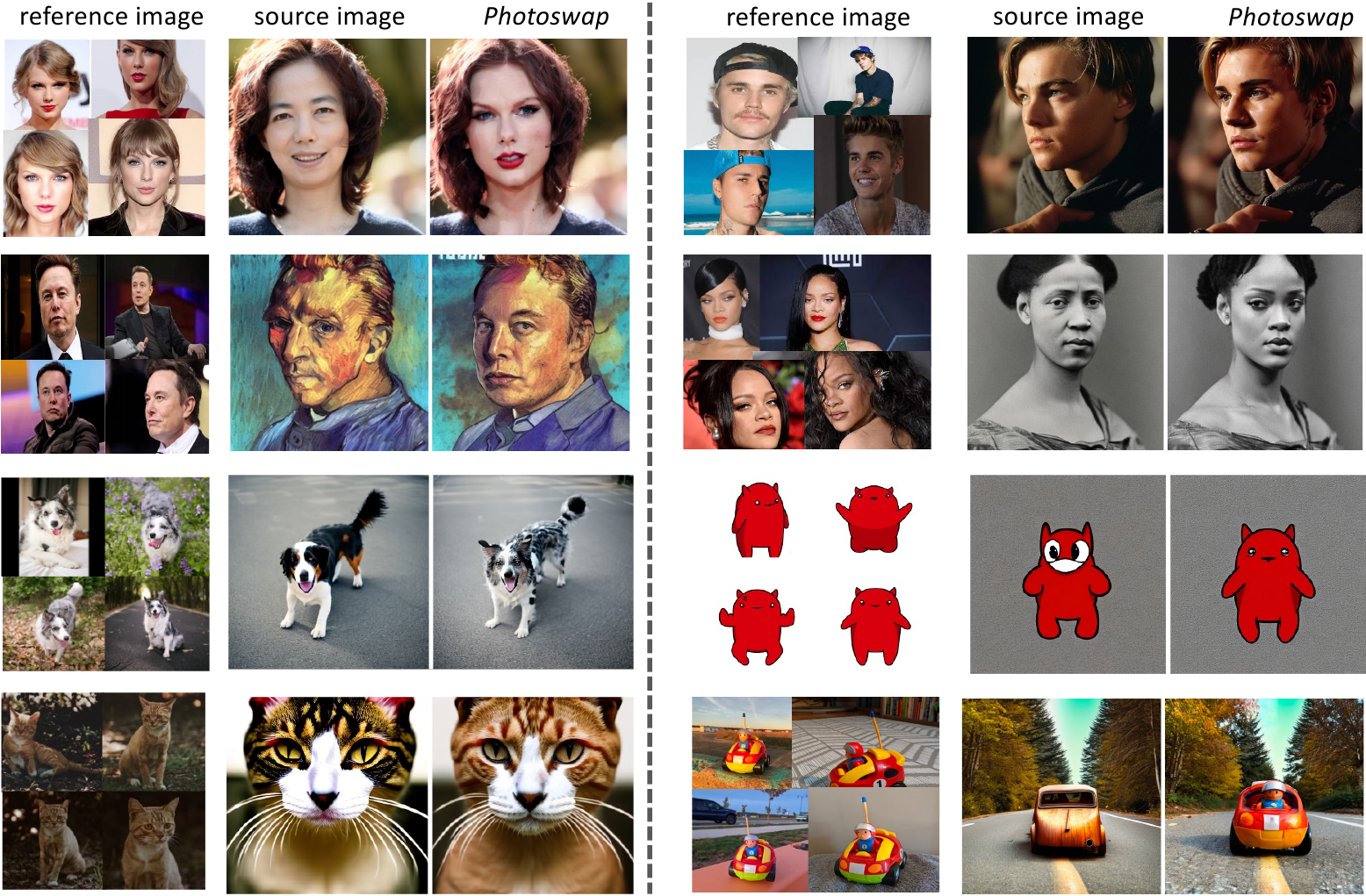}
    \caption{\textbf{\emph{Photoswap} results across various object and image domains, demonstrating its wide applicability.} From everyday objects to cartoon, the diversity in subject swapping tasks has showcased the versatility and robustness of our framework across different contexts.
    }
    \label{fig:result_teaser}
\end{figure}

\section{Experiments}
\subsection{Imlementation Details}
\label{sec:implementation}
For the implementation of subject swapping on real images, we require an additional process that utilizes an image inversion method, specifically the DDIM inversion~\cite{song2020denoising}, to transform the image into initial noise. This inversion method relies on a reversed sequence of sampling to achieve the desired inversion.
However, there exist inherent challenges when this inversion process is applied in text-guided synthesis within a classifier-free guidance setting. Notably, the inversion can potentially amplify the accumulated error, which could ultimately lead to subpar reconstruction outcomes.
To fortify the robustness of the DDIM inversion and to mitigate this issue, we further optimize the null text embedding, as detailed in~\citet{mokady2022null-text}. The incorporation of this optimization technique bolsters the effectiveness and reliability of the inversion process, consequently allowing for a more precise reconstruction. Without further notice, the DDIM inversion in this paper is enhanced by null text embedding optimization.

During inference, we utilize the DDIM sampling method with 50 denoising steps and classifier-free guidance of 7.5. The default step $\lambda_A$ for cross-attention map replacement is 20. The default step $\lambda_M$ for self-attention map replacement is 25, while the default step for self-attention feature $\lambda_\phi$ replacement is 10. Note that the replacement steps may change to some specific checkpoint. As mentioned in Section~\ref{sec:method}, the target prompt $P_t$ is just source prompt $P_s$ with the object token being replaced with the new concept token.  For concept learning, we mainly utilize DreamBooth~\cite{ruiz2023dream-booth} to finetune a stable diffusion 2.1 to learn the new concept from 3~5 images. The learning rate is set to 1e-6. We use Adawm optimizer with 800 hundred training step. We finetune both the U-net and text encoder. The DreamBooth training takes around 10 minutes on a machine with 8 A100 GPU cards.

\subsection{Personalized Subject Swapping Results}

Figure ~\ref{fig:result_teaser} showcases the effectiveness of our \emph{Photoswap} technique for subject swapping. Our approach excels at preserving crucial aspects such as spatial layout, geometry, and the pose of the original subject while seamlessly introducing a reference subject into the target image. Remarkably, even in cartoon images, our method ensures that the background remains intact during the subject change process. A notable example is the "cat" image, where our technique successfully retains all the intricate details from the source image, including the distinctive "Whiskers." This demonstrates our framework's ability to accurately capture and preserve fine-grained information during subject swapping.  
\begin{figure}
     \centering
     \begin{subfigure}[b]{0.47\textwidth}
         \centering
         \includegraphics[width=\textwidth]{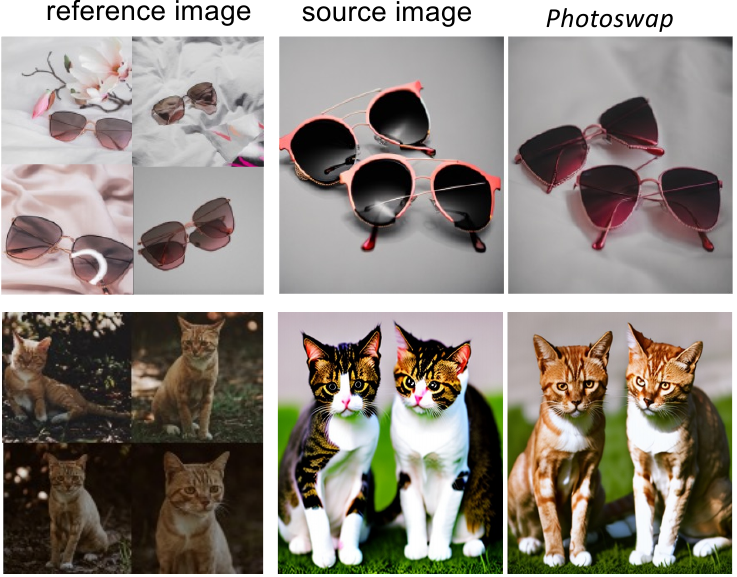}
         \caption{Multi-subject swap.}
         \label{fig:multiple-subjects}
     \end{subfigure}
     \hfill
     \begin{subfigure}[b]{0.47\textwidth}
         \centering
         \includegraphics[width=\textwidth]{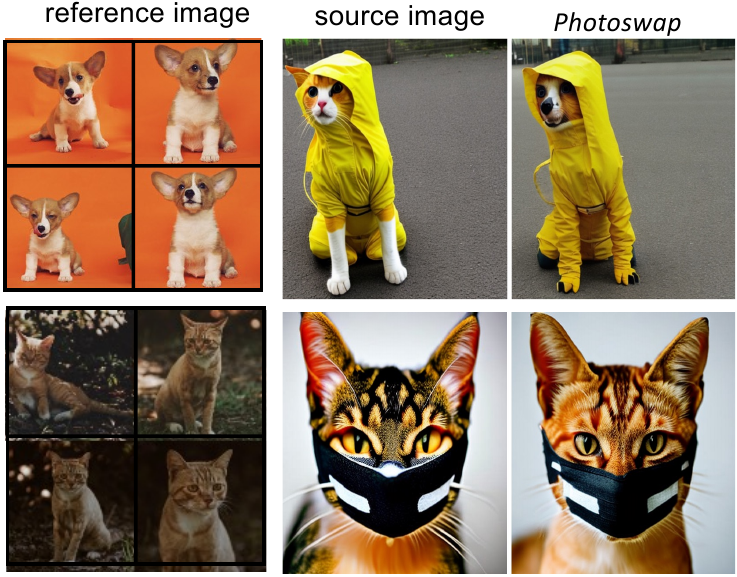}
         \caption{Occluded subject swap.}
         \label{fig:occluded}
     \end{subfigure}
     \caption{\textbf{\emph{Photoswap} results on multi-subject and occluded subject scenarios.} The results show that \emph{Photoswap} can disentangle and replace multiple subjects at once. Also, \emph{Photoswap} can identify the target object while avoiding influencing the non-subject pixels.}
        \label{fig:multiple-and-occluded}
\end{figure}

We further demonstrate the versatility of \emph{Photoswap} by showcasing its effectiveness in multiple subject swap and occluded object swap scenarios. As depicted in Figure~\ref{fig:real_three_faces} (a), we present a source image featuring two sunglasses, which are successfully replaced with reference glass while preserving the original layout of the sunglasses. Similarly, in Figure~\ref{fig:real_three_faces} (b), we observe a source image with a dog partially occluded by a suit. The resulting swapped dog wears a suit that closely matches the occluded region. These examples serve to highlight the robustness of our proposed \emph{Photoswap} method in handling various real-world cases, thereby enabling users to explore a broader range of editing possibilities.

\subsection{Comparison with Baseline Methods}
\begin{figure}
    \centering
    \includegraphics[width=0.7\textwidth]{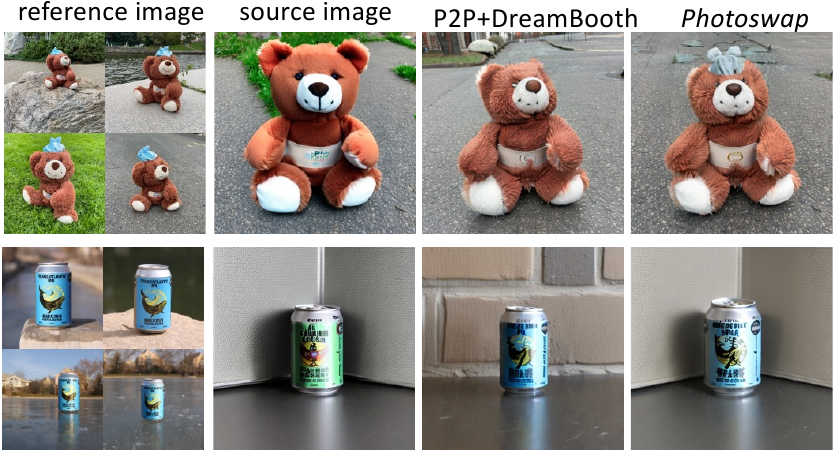}
    \caption{\textbf{Qualitative comparison between P2P+DreamBooth and \emph{Photoswap}.} We can observe that P2P+dreambooth is capable of achieving subject swapping. However, it faces challenges in preserving both the background and the reference subject accurately, while for \emph{Photoswap}, it is robust to handle various cases.
    }
    \label{fig:gesture-modification}
\end{figure}

Personalized object swap is a new task and there is no existing benchmark. However, we could modify the existing attention manipulation based methods. More specifically, we used the same concept learning method DreamBooth to finetune the same stable diffusion checkpoint to inject the new concept. To fairly compare with our results, we modified existing prompt-based editing method P2P~\cite{hertz2022prompt} ,  an editing method based diffusion models.  Note that origin P2P only works on a pair of synthetic images, in our setting we use same concept learning dreambooth an fix the seed to allow concept swapping.  On the other hand, PnP~\cite{tumanyan2022plug} could also be implmented in similar setting, however we found PnP usually can not lead to satisfactory object swapping and may lead to a huge difference between the source image and the generated image. We suspect that it is because PnP is designed for image translation so it does not initiate the attention manipulation step from the beginning step. The qualitative comparision between \emph{Photoswap} and P2P+dreambooth is shown in Figure ~\ref{fig:gesture-modification}.
We observe that P2P with DreamBooth could achieve achieve basic object swap, but it still suffers from background mismatching issue.

\begin{table}[t]
\centering
    \begin{tabular}{cccc}
        \toprule
        ~ & \textit{Photoswap} & P2P+DreamBooth & Tie \\
        \midrule
        Subject Swapping ~ & \textbf{46.8\%} & 25.6\% & 27.6\% \\
        Background Preservation ~ & \textbf{40.7\%} & 32.7\% & 26.6\% \\
        Overall Quality & \textbf{50.8\%} & 28.0\% & 21.2\% \\
        \bottomrule
    \end{tabular}
    \vspace{1ex}
    \caption{\textbf{Human evaluation between \emph{Photoswap} and P2P}. Note that both \emph{Photoswap} and P2P leverage same concept learning method DreamBooth. The results indicates, for most of the cases above 70\%, the proposed \emph{Photoswap} is better or on par with P2P.
    }
    \label{table:human}
\end{table}
\paragraph{Human Evaluation.}
We conduct a human evaluation to study the editing quality by (1) \textit{Which result better swaps the subject as the reference and keeps its identity}; (2) \textit{Which result better preserves the background}; (3) \textit{Which result has better overall subject-driven swapping}. We randomly sample 99 examples and adopt Amazon MTurk\footnote{Amazon Mechanical Turk (MTurk): \url{https://www.mturk.com.}} to compare between two results. To avoid potential bias, we hire 3 Turkers for each sample. Table~\ref{table:human} demonstrates the comparison between our \emph{Photoswap} and P2P. Firstly, more turkers (over 46\%) denote that our \emph{Photoswap} better swaps the subject yet keeps its identity at the same time. Moreover, we can also preserve the background in the source image (41\% \textit{vs.} 33\%), which is another crucial goal of this editing. In summary, \emph{Photoswap} precisely performs subject swapping and preserves the remaining part from the input, leading to an overall superiority (50\%) to P2P.

\subsection{Controlling Subject Identity}
\begin{figure}
    \centering
    \includegraphics[width=\textwidth]{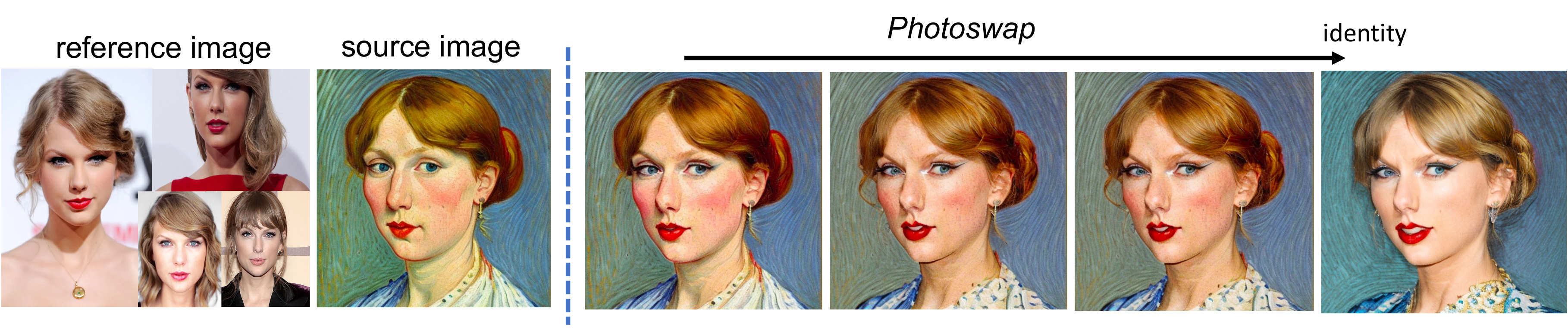}
    \caption{\textbf{Ablation results of $M$ in \emph{Photoswap}.} With $M$ value increase, the generated one is more similar to the style and identity of source image and dissimilar to the reference subject,  vice versa.}
    \label{fig:result_ablation}
\end{figure}

The effectiveness of the proposed mutual self-attention is demonstrated through both synthetic image synthesis and real image editing. Additionally, we perform an analysis of the control strategy with varying values of $M$ during the denoising process. Figure ~\ref{fig:result_ablation} provides insights into this analysis. It is observed that when applying self-attention control with a large swapping step $\lambda_M$ for $M$, the synthesized image closely resembles the source image in terms of both style and identity. In this scenario, all contents from the source image are preserved, while the subject style learned from the reference subject is disregarded. As the value of $M$ decreases, the synthesized image maintains the subject from the reference image while retaining the layout and pose of the contents from the source image. This gradual transition in the control strategy allows for a balance between subject style transfer and preservation of the original image's contents.

\subsection{Attention Swapping Step Analysis}
In this section, we visualize the effect of the influence of swapping steps of different components. 
As discussed in the main paper, self-attention output \textbf{$\phi$}, and self-attention map \textbf{$M$}, derived from the self-attention layer, encompasses comprehensive content information from the source image, without relying on direct computation with textual features. 
Previous works such as~\citet{hertz2022prompt} did not explore the usage of \textbf{$\phi$} and \textbf{$M$} in the object-level image editing process. 

\begin{figure}[!th]
    \centering
    \includegraphics[width=\textwidth]{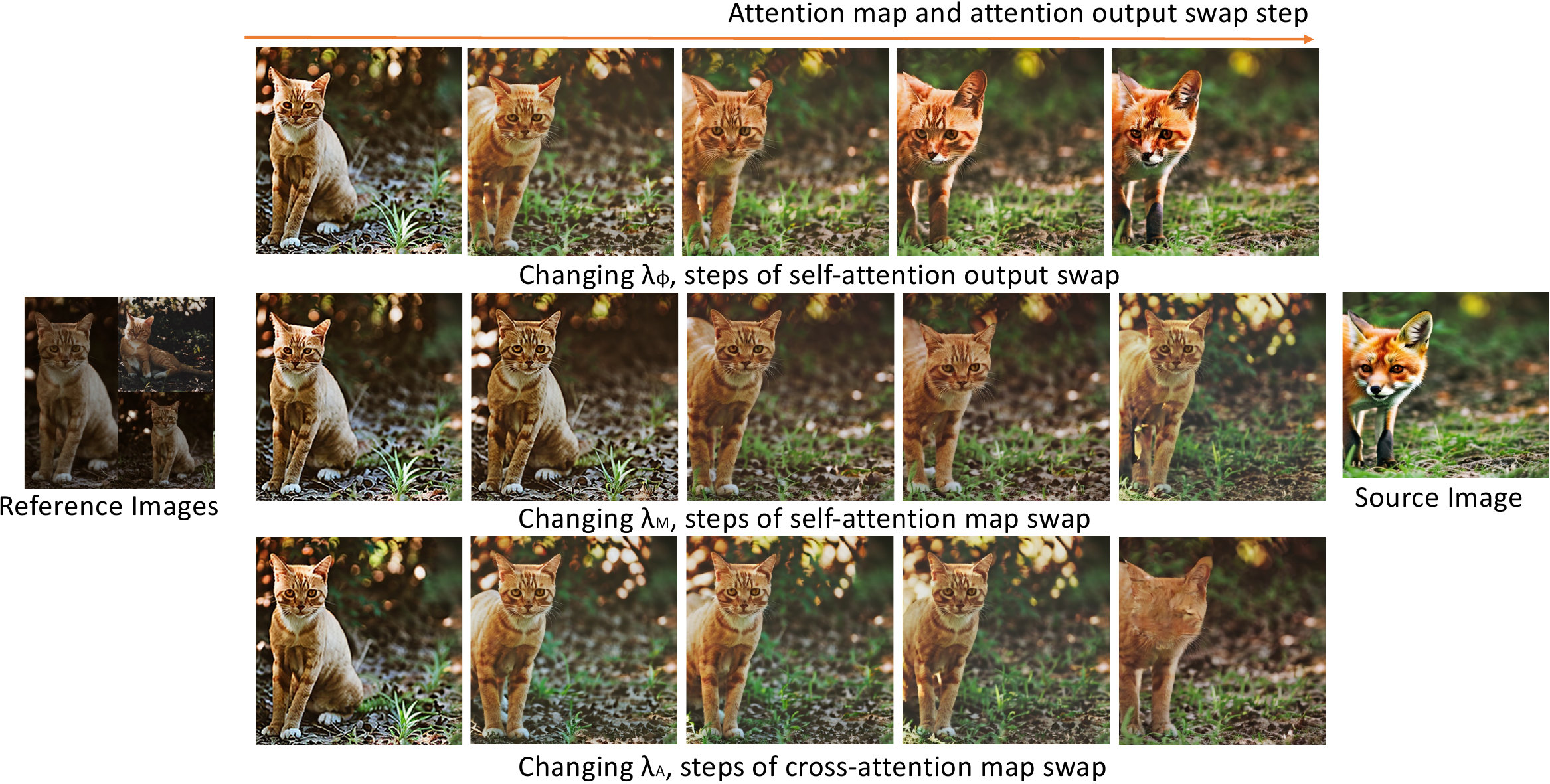}
    \caption{\textbf{Results at different swapping steps}. With consistent steps, swapping the self-attention output provides superior control over the layout, including the subject's gestures and the background details. 
    However, excessive swapping could affect the subject's identity, as the new concept introduced through the text prompt might be overshadowed by the swapping of the attention output or attention map. 
    This effect is more clear when swapping the self-attention output $\lambda_\phi$. 
    Furthermore, we observed that replacing the attention map for an extensive number of steps can result in an image with significant noise, possibly due to a compatibility issue between the attention map and the $v$ vector.
    }
    \label{fig:step-analysis}
\end{figure}

Figure~\ref{fig:step-analysis} provides a visual representation of the effect of incrementally increasing the swapping step for one $\lambda$ hyperparameter while maintaining the other two at zero. Although all of them can be utilized for subject swapping, they demonstrate varying levels of layout control.
At the same swapping step, the self-attention output \textbf{$\phi$} offers more robust layout control, facilitating better alignment of gestures and preservation of background context. In contrast, the self-attention map $\boldsymbol{M}$ and cross-attention map $\boldsymbol{A}$ demonstrate similar capabilities in controlling the layout.

However, extensive swapping can affect the subject's identity, as the novel concept introduced via the text prompt might be eclipsed by the swapping of the attention output or attention map. This effect becomes particularly evident when swapping the self-attention output. This analysis further informs the determination of the default $\lambda_\phi$, $\lambda_M$, and $\lambda_A$ values.
While the cross-attention map $\boldsymbol{A}$ facilitates more fine-grained generation control, given its incorporation of information from textual tokens, we discovered that $\boldsymbol{\phi}$ offers stronger holistic generation control, bolstering the overall output's quality and integrity.


\subsection{Results of Other Concept Learning Methods}
\label{sec:appendix:other_concept_learning}


Our mainly use DreamBooth as the concept learning method in the experiments, primarily due to its superior capabilities in learning subject identities~\cite{ruiz2023dream-booth}. However, our method is not strictly dependent on any specific concept learning method. In fact, other concept learning methods could be effectively employed to introduce the concept of the target subject.

\begin{figure}[!th]
    \centering
    \includegraphics[width=\textwidth]{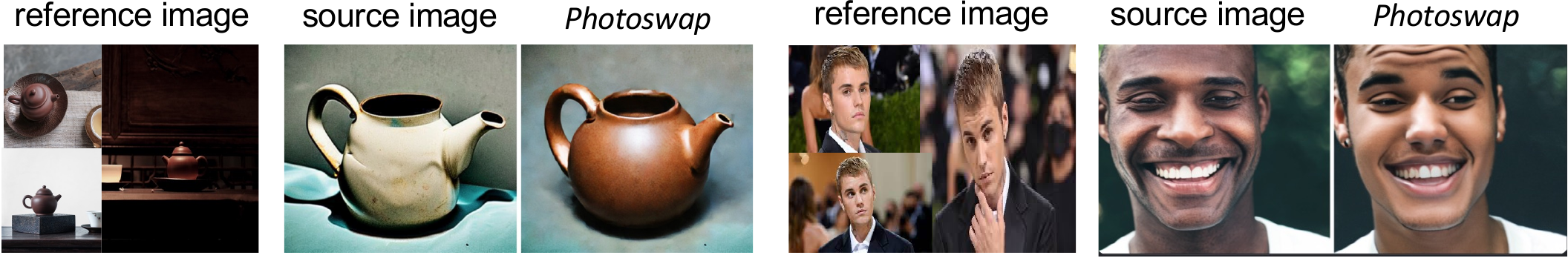}
    \caption{\textbf{Results of Text Inversion~\cite{gal2023ti} as the concept learning module}. It can successfully capture key subject features, but its performance drops when representing complex structures such as human faces.}
    \label{fig:text-inversion}
\end{figure}

To illustrate this, we present the results of \modelname{} when applying Text Inversion~\cite{gal2023ti}. We train the model using 8 A100 GPUs with a batch size of 4, a learning rate of 5e-4, and set the training steps to 1000. Results in Figure~\ref{fig:text-inversion} indicate that Text Inversion also proves to be an effective concept learning method, as it successfully captures key features of the target object.
Nevertheless, we observe that Text Inversion performance is notably underwhelming when applied to human faces. We postulate that this is because Text Inversion focuses on learning a new embedding for the novel concept, rather than finetuning the entire model. Consequently, the capacity to express the new concept becomes inherently limited, resulting in its less than optimal performance in certain areas. 
\subsection{Ethics Exploration}

\begin{figure}[!th]
    \centering
    \includegraphics[width=1.0\textwidth]{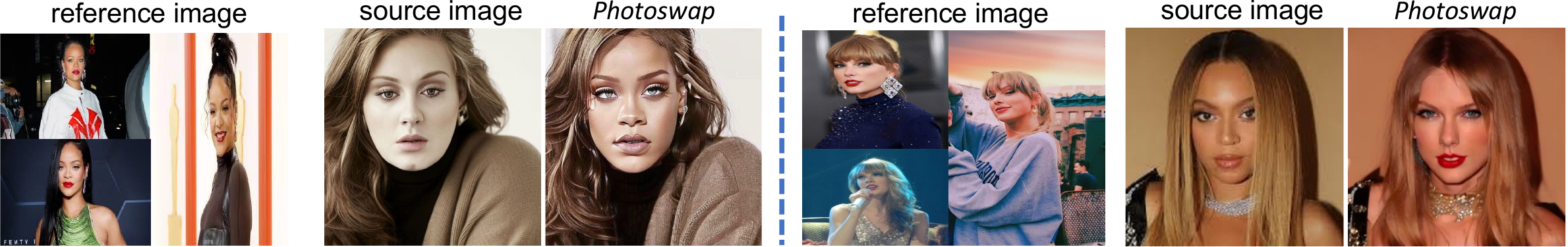}
    \caption{\textbf{Results on real human face images across different races}. Evidently, the skin colors are also successfully transferred when swapping a white person with a black person, and vice versa.
    }
    \label{fig:ethics}
\end{figure}

Like many AI technologies, text-to-image diffusion models can potentially exhibit biases reflective of those inherent in the training data~\cite{sasha2023stable, perera2023analyzing}. Given that these models are trained on vast text and image datasets, they might inadvertently learn and perpetuate biases, such as stereotypes and prejudices, found within this data. For instance, should the training data contain skewed representations or descriptions of specific demographic groups, the model may produce biased images in response to related prompts.

However, \modelname{} has been designed to mitigate bias within the generation process of a text-to-image diffusion model. It achieves this by directly substituting the depicted subject with the intended target.
In Figure ~\ref{fig:ethics}, we present our evaluation of face swapping across various skin tones. It is crucial to observe that when there is a significant disparity between the source and reference images, the swapping results tend to homogenize the skin color. 
As a result, we advocate for the use of \modelname{} on subjects of similar racial backgrounds to achieve more satisfactory and authentic outcomes. Despite these potential disparities, the model ensures the preservation of most of the target subject's specific facial features, reinforcing the credibility and accuracy of the final image.


\subsection{Failure Cases}
\label{sec:failure-cases}

\begin{figure}[!ht]
    \centering
    \includegraphics[width=1.0\textwidth]{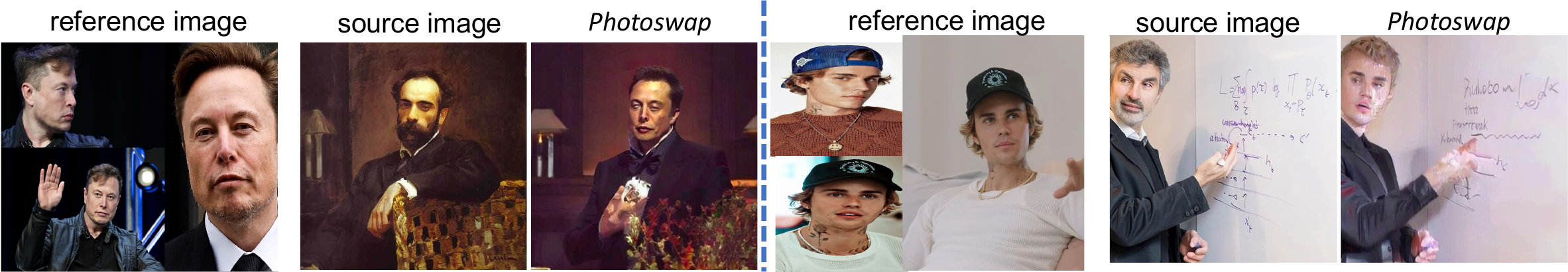}
    \caption{\textbf{Failure cases.} The model sometimes struggles to accurately reconstruct hand details and complex background information such as formula on a whiteboard.}
    \label{fig:failure_cases}
\end{figure}

Here we highlight two common failure cases. First, the model struggles to accurately reproduce hands. When the subject includes hands and fingers, the swapping results often fail to precisely mirror the original hand gestures or the number of fingers. This issue could be an inherited challenge from Stable Diffusion. 
Moreover, \modelname{} can encounter difficulties when the image comprises complex information. As illustrated in the lower row of Figure~\ref{fig:failure_cases}, \modelname{} fails to reconstruct the complicated formula on a whiteboard. 
Therefore, while \modelname{} exhibits strong performance across various scenarios, it's crucial to acknowledge these limitations when considering its application in real-world scenarios involving intricate hand gestures or complex abstract information.

\section{Conclusion}
\label{sec:conclusion}
This paper introduces \emph{Photoswap}, a novel framework designed for personalized subject swapping in images. To facilitate seamless subject photo swapping, we propose leveraging self-attention control by exchanging intermediate variables within the attention layer between the source image and reference images. Despite its simplicity, our extensive experimentation and evaluations provide compelling evidence for the effectiveness of \emph{Photoswap}. Our framework offers a robust and intuitive solution for subject swapping, enabling users to effortlessly manipulate images according to their preferences. In the future, we plan to further advance the method to address those common failure issues to enhance the overall performance and versatility of personalized subject swapping.

{\small
\bibliographystyle{natbib}
\bibliography{reference}

\begin{thebibliography}{}

\bibitem[Blattmann {\em et~al.}(2022)Blattmann, Rombach, Oktay, Müller, and
  Ommer]{blattmann2022retrieval-diffusion}
Blattmann, A., Rombach, R., Oktay, K., Müller, J., and Ommer, B. (2022).
\newblock {Retrieval-Augmented Diffusion Models}.
\newblock In {\em NeurIPS\/}.

\bibitem[Brock {\em et~al.}(2018)Brock, Donahue, and Simonyan]{brock2018large}
Brock, A., Donahue, J., and Simonyan, K. (2018).
\newblock Large scale gan training for high fidelity natural image synthesis.
\newblock {\em arXiv\/}.

\bibitem[Cao {\em et~al.}(2023)Cao, Wang, Qi, Shan, Qie, and
  Zheng]{cao2023masactrl}
Cao, M., Wang, X., Qi, Z., Shan, Y., Qie, X., and Zheng, Y. (2023).
\newblock Masactrl: Tuning-free mutual self-attention control for consistent
  image synthesis and editing.
\newblock {\em arXiv\/}.

\bibitem[Casanova {\em et~al.}(2021)Casanova, Careil, Verbeek, Drozdzal, and
  Romero-Soriano]{casanova2021ic-gan}
Casanova, A., Careil, M., Verbeek, J., Drozdzal, M., and Romero-Soriano, A.
  (2021).
\newblock {Instance-Conditioned GAN}.
\newblock In {\em NeurIPS\/}.

\bibitem[Chen {\em et~al.}(2023)Chen, Hu, Saharia, and
  Cohen]{chen2023re-imagen}
Chen, W., Hu, H., Saharia, C., and Cohen, W.~W. (2023).
\newblock {Re-Imagen: Retrieval-Augmented Text-to-Image Generator}.
\newblock In {\em ICLR\/}.

\bibitem[Couairon {\em et~al.}(2022)Couairon, Verbeek, Schwenk, and
  Cord]{couairon2022diffedit}
Couairon, G., Verbeek, J., Schwenk, H., and Cord, M. (2022).
\newblock Diffedit: Diffusion-based semantic image editing with mask guidance.
\newblock {\em arXiv\/}.

\bibitem[Crowson {\em et~al.}(2022)Crowson, Biderman, Kornis, Stander,
  Hallahan, Castricato, and Raff]{crowson2022vqgan}
Crowson, K., Biderman, S., Kornis, D., Stander, D., Hallahan, E., Castricato,
  L., and Raff, E. (2022).
\newblock Vqgan-clip: Open domain image generation and editing with natural
  language guidance.
\newblock In {\em ECCV\/}.

\bibitem[Deng {\em et~al.}(2022)Deng, Tang, Dong, Ma, Pan, Wang, and
  Xu]{deng2022stytr2}
Deng, Y., Tang, F., Dong, W., Ma, C., Pan, X., Wang, L., and Xu, C. (2022).
\newblock Stytr2: Image style transfer with transformers.
\newblock In {\em CVPR\/}.

\bibitem[Ding {\em et~al.}(2021)Ding, Yang, Hong, Zheng, Zhou, Yin, Lin, Zou,
  Shao, Yang, {\em et~al.}]{ding2021cogview}
Ding, M., Yang, Z., Hong, W., Zheng, W., Zhou, C., Yin, D., Lin, J., Zou, X.,
  Shao, Z., Yang, H., {\em et~al.} (2021).
\newblock Cogview: Mastering text-to-image generation via transformers.
\newblock {\em NeurIPS\/}.

\bibitem[Feng {\em et~al.}(2023)Feng, He, Fu, Jampani, Akula, Narayana, Basu,
  Wang, and Wang]{feng2023training-free}
Feng, W., He, X., Fu, T.-J., Jampani, V., Akula, A., Narayana, P., Basu, S.,
  Wang, X.~E., and Wang, W.~Y. (2023).
\newblock {Training-Free Structured Diffusion Guidance for Compositional
  Text-to-Image Synthesis}.
\newblock In {\em ICLR\/}.

\bibitem[Gal {\em et~al.}(2022)Gal, Alaluf, Atzmon, Patashnik, Bermano,
  Chechik, and Cohen-Or]{gal2022image}
Gal, R., Alaluf, Y., Atzmon, Y., Patashnik, O., Bermano, A.~H., Chechik, G.,
  and Cohen-Or, D. (2022).
\newblock An image is worth one word: Personalizing text-to-image generation
  using textual inversion.
\newblock {\em arXiv\/}.

\bibitem[Gal {\em et~al.}(2023a)Gal, Alaluf, Atzmon, Patashnik, Bermano,
  Chechik, and Cohen-Or]{gal2023ti}
Gal, R., Alaluf, Y., Atzmon, Y., Patashnik, O., Bermano, A.~H., Chechik, G.,
  and Cohen-Or, D. (2023a).
\newblock {An Image is Worth One Word: Personalizing Text-to-Image Generation
  using Textual Inversion}.
\newblock In {\em ICLR\/}.

\bibitem[Gal {\em et~al.}(2023b)Gal, Arar, Atzmon, Bermano, Chechik, and
  Cohen-Or]{gal2023edt}
Gal, R., Arar, M., Atzmon, Y., Bermano, A.~H., Chechik, G., and Cohen-Or, D.
  (2023b).
\newblock {Encoder-based Domain Tuning for Fast Personalization of
  Text-to-Image Models}.
\newblock In {\em arXiv\/}.

\bibitem[Goodfellow {\em et~al.}(2020)Goodfellow, Pouget-Abadie, Mirza, Xu,
  Warde-Farley, Ozair, Courville, and Bengio]{goodfellow2020generative}
Goodfellow, I., Pouget-Abadie, J., Mirza, M., Xu, B., Warde-Farley, D., Ozair,
  S., Courville, A., and Bengio, Y. (2020).
\newblock Generative adversarial networks.
\newblock {\em Communications of the ACM\/}.

\bibitem[Gu {\em et~al.}(2022)Gu, Chen, Bao, Wen, Zhang, Chen, Yuan, and
  Guo]{gu2022vector}
Gu, S., Chen, D., Bao, J., Wen, F., Zhang, B., Chen, D., Yuan, L., and Guo, B.
  (2022).
\newblock Vector quantized diffusion model for text-to-image synthesis.
\newblock In {\em CVPR\/}.

\bibitem[Hertz {\em et~al.}(2022)Hertz, Mokady, Tenenbaum, Aberman, Pritch, and
  Cohen-Or]{hertz2022prompt}
Hertz, A., Mokady, R., Tenenbaum, J., Aberman, K., Pritch, Y., and Cohen-Or, D.
  (2022).
\newblock Prompt-to-prompt image editing with cross attention control.
\newblock {\em arXiv\/}.

\bibitem[Huang {\em et~al.}(2018)Huang, Liu, Belongie, and
  Kautz]{huang2018multimodal}
Huang, X., Liu, M.-Y., Belongie, S., and Kautz, J. (2018).
\newblock Multimodal unsupervised image-to-image translation.
\newblock In {\em ECCV\/}.

\bibitem[Jahn {\em et~al.}(2021)Jahn, Rombach, and Ommer]{jahn2021high}
Jahn, M., Rombach, R., and Ommer, B. (2021).
\newblock High-resolution complex scene synthesis with transformers.
\newblock {\em arXiv\/}.

\bibitem[Karras {\em et~al.}(2019)Karras, Laine, and Aila]{karras2019style}
Karras, T., Laine, S., and Aila, T. (2019).
\newblock A style-based generator architecture for generative adversarial
  networks.
\newblock In {\em CVPR\/}.

\bibitem[Kawar {\em et~al.}(2022)Kawar, Zada, Lang, Tov, Chang, Dekel, Mosseri,
  and Irani]{kawar2022imagic}
Kawar, B., Zada, S., Lang, O., Tov, O., Chang, H., Dekel, T., Mosseri, I., and
  Irani, M. (2022).
\newblock Imagic: Text-based real image editing with diffusion models.
\newblock {\em arXiv\/}.

\bibitem[Kumari {\em et~al.}(2023)Kumari, Zhang, Zhang, Shechtman, and
  Zhu]{kumari2023mcc}
Kumari, N., Zhang, B., Zhang, R., Shechtman, E., and Zhu, J.-Y. (2023).
\newblock {Multi-Concept Customization of Text-to-Image Diffusion}.
\newblock In {\em CVPR\/}.

\bibitem[Li {\em et~al.}(2023)Li, Liu, Wu, Mu, Yang, Gao, Li, and
  Lee]{li2023gligen}
Li, Y., Liu, H., Wu, Q., Mu, F., Yang, J., Gao, J., Li, C., and Lee, Y.~J.
  (2023).
\newblock Gligen: Open-set grounded text-to-image generation.
\newblock {\em arXiv\/}.

\bibitem[Liao {\em et~al.}(2017)Liao, Yao, Yuan, Hua, and Kang]{liao2017visual}
Liao, J., Yao, Y., Yuan, L., Hua, G., and Kang, S.~B. (2017).
\newblock Visual atribute transfer through deep image analogy.
\newblock {\em ACM Transactions on Graphics\/}.

\bibitem[Liu {\em et~al.}(2021)Liu, Lin, He, Li, Wang, Li, Sun, Li, and
  Ding]{liu2021adaattn}
Liu, S., Lin, T., He, D., Li, F., Wang, M., Li, X., Sun, Z., Li, Q., and Ding,
  E. (2021).
\newblock Adaattn: Revisit attention mechanism in arbitrary neural style
  transfer.
\newblock In {\em ICCV\/}.

\bibitem[Meng {\em et~al.}(2021)Meng, Song, Song, Wu, Zhu, and
  Ermon]{meng2021sdedit}
Meng, C., Song, Y., Song, J., Wu, J., Zhu, J.-Y., and Ermon, S. (2021).
\newblock Sdedit: Image synthesis and editing with stochastic differential
  equations.
\newblock {\em arXiv\/}.

\bibitem[Mokady {\em et~al.}(2022)Mokady, Hertz, Aberman, Pritch, and
  Cohen-Or]{mokady2022null-text}
Mokady, R., Hertz, A., Aberman, K., Pritch, Y., and Cohen-Or, D. (2022).
\newblock {Null-text Inversion for Editing Real Images using Guided Diffusion
  Models}.
\newblock In {\em arXiv\/}.

\bibitem[Nichol {\em et~al.}(2021)Nichol, Dhariwal, Ramesh, Shyam, Mishkin,
  McGrew, Sutskever, and Chen]{nichol2021glide}
Nichol, A., Dhariwal, P., Ramesh, A., Shyam, P., Mishkin, P., McGrew, B.,
  Sutskever, I., and Chen, M. (2021).
\newblock Glide: Towards photorealistic image generation and editing with
  text-guided diffusion models.
\newblock {\em arXiv\/}.

\bibitem[Nitzan {\em et~al.}(2022)Nitzan, Aberman, He, Liba, Yarom, Gandelsman,
  Mosseri, Pritch, and Cohen-or]{nitzan2022my-style}
Nitzan, Y., Aberman, K., He, Q., Liba, O., Yarom, M., Gandelsman, Y., Mosseri,
  I., Pritch, Y., and Cohen-or, D. (2022).
\newblock {MyStyle: A Personalized Generative Prior}.
\newblock In {\em Special Interest Group on Computer Graphics and Interactive
  Techniques in Asia (SIGGRAPH Asia)\/}.

\bibitem[OpenAI(2021)OpenAI]{dalle}
OpenAI (2021).
\newblock {DALL·E: Creating images from text}.
\newblock \url{https://openai.com/research/dall-e}.

\bibitem[OpenAI(2022)OpenAI]{dalle2}
OpenAI (2022).
\newblock {DALL·E2}.
\newblock \url{https://openai.com/product/dall-e-2}.

\bibitem[Perera and Patel(2023)Perera and Patel]{perera2023analyzing}
Perera, M.~V. and Patel, V.~M. (2023).
\newblock Analyzing bias in diffusion-based face generation models.
\newblock {\em arXiv preprint arXiv:2305.06402\/}.

\bibitem[Radford {\em et~al.}(2021)Radford, Kim, Hallacy, Ramesh, Goh, Agarwal,
  Sastry, Askell, Mishkin, Clark, {\em et~al.}]{radford2021learning}
Radford, A., Kim, J.~W., Hallacy, C., Ramesh, A., Goh, G., Agarwal, S., Sastry,
  G., Askell, A., Mishkin, P., Clark, J., {\em et~al.} (2021).
\newblock Learning transferable visual models from natural language
  supervision.
\newblock In {\em ICML\/}.

\bibitem[Rombach {\em et~al.}(2022)Rombach, Blattmann, Lorenz, Esser, and
  Ommer]{rombach2022high}
Rombach, R., Blattmann, A., Lorenz, D., Esser, P., and Ommer, B. (2022).
\newblock High-resolution image synthesis with latent diffusion models.
\newblock In {\em CVPR\/}.

\bibitem[Ronneberger {\em et~al.}(2015)Ronneberger, Fischer, and
  Brox]{ronneberger2015u}
Ronneberger, O., Fischer, P., and Brox, T. (2015).
\newblock U-net: Convolutional networks for biomedical image segmentation.
\newblock In {\em MICCAI\/}. Springer.

\bibitem[Ruiz {\em et~al.}(2023)Ruiz, Li, Jampani, Pritch, Rubinstein, and
  Aberman]{ruiz2023dream-booth}
Ruiz, N., Li, Y., Jampani, V., Pritch, Y., Rubinstein, M., and Aberman, K.
  (2023).
\newblock {DreamBooth: Fine Tuning Text-to-Image Diffusion Models for
  Subject-Driven Generation}.
\newblock In {\em CVPR\/}.

\bibitem[Saharia {\em et~al.}(2022)Saharia, Chan, Saxena, Li, Whang, Denton,
  Ghasemipour, Gontijo~Lopes, Karagol~Ayan, Salimans, {\em
  et~al.}]{saharia2022photorealistic}
Saharia, C., Chan, W., Saxena, S., Li, L., Whang, J., Denton, E.~L.,
  Ghasemipour, K., Gontijo~Lopes, R., Karagol~Ayan, B., Salimans, T., {\em
  et~al.} (2022).
\newblock Photorealistic text-to-image diffusion models with deep language
  understanding.
\newblock In {\em NeurIPS\/}.

\bibitem[Sasha~Luccioni {\em et~al.}(2023)Sasha~Luccioni, Akiki, Mitchell, and
  Jernite]{sasha2023stable}
Sasha~Luccioni, A., Akiki, C., Mitchell, M., and Jernite, Y. (2023).
\newblock Stable bias: Analyzing societal representations in diffusion models.
\newblock {\em arXiv e-prints\/}, pages arXiv--2303.

\bibitem[Seo {\em et~al.}(2022)Seo, Lee, Cho, Lee, and Kim]{seo2022midms}
Seo, J., Lee, G., Cho, S., Lee, J., and Kim, S. (2022).
\newblock Midms: Matching interleaved diffusion models for exemplar-based image
  translation.
\newblock {\em arXiv\/}.

\bibitem[Sheynin {\em et~al.}(2023)Sheynin, Ashual, Polyak, Singer, Gafni,
  Nachmani, and Taigman]{sheynin2023knn-diffusion}
Sheynin, S., Ashual, O., Polyak, A., Singer, U., Gafni, O., Nachmani, E., and
  Taigman, Y. (2023).
\newblock {KNN-Diffusion: Image Generation via Large-Scale Retrieval}.
\newblock In {\em ICLR\/}.

\bibitem[Song {\em et~al.}(2020)Song, Meng, and Ermon]{song2020denoising}
Song, J., Meng, C., and Ermon, S. (2020).
\newblock Denoising diffusion implicit models.
\newblock In {\em International Conference on Learning Representations\/}.

\bibitem[Tumanyan {\em et~al.}(2022)Tumanyan, Geyer, Bagon, and
  Dekel]{tumanyan2022plug}
Tumanyan, N., Geyer, M., Bagon, S., and Dekel, T. (2022).
\newblock Plug-and-play diffusion features for text-driven image-to-image
  translation.
\newblock {\em arXiv\/}.

\bibitem[Wang {\em et~al.}(2019)Wang, Yang, Li, Liang, Zhang, Hall, and
  Hu]{wang2019example}
Wang, M., Yang, G.-Y., Li, R., Liang, R.-Z., Zhang, S.-H., Hall, P.~M., and Hu,
  S.-M. (2019).
\newblock Example-guided style-consistent image synthesis from semantic
  labeling.
\newblock In {\em CVPR\/}.

\bibitem[Yang {\em et~al.}(2022a)Yang, Gu, Zhang, Zhang, Chen, Sun, Chen, and
  Wen]{yang2022paint}
Yang, B., Gu, S., Zhang, B., Zhang, T., Chen, X., Sun, X., Chen, D., and Wen,
  F. (2022a).
\newblock Paint by example: Exemplar-based image editing with diffusion models.
\newblock {\em arXiv\/}.

\bibitem[Yang {\em et~al.}(2022b)Yang, Wang, Gan, Li, Lin, Wu, Duan, Liu, Liu,
  Zeng, {\em et~al.}]{yang2022reco}
Yang, Z., Wang, J., Gan, Z., Li, L., Lin, K., Wu, C., Duan, N., Liu, Z., Liu,
  C., Zeng, M., {\em et~al.} (2022b).
\newblock Reco: Region-controlled text-to-image generation.
\newblock {\em arXiv\/}.

\bibitem[Zeng {\em et~al.}(2022)Zeng, Lin, Zhang, Liu, Collomosse, Kuen, and
  Patel]{zeng2022scenecomposer}
Zeng, Y., Lin, Z., Zhang, J., Liu, Q., Collomosse, J., Kuen, J., and Patel,
  V.~M. (2022).
\newblock Scenecomposer: Any-level semantic image synthesis.
\newblock {\em arXiv\/}.

\bibitem[Zhang and Agrawala(2023)Zhang and Agrawala]{zhang2023adding}
Zhang, L. and Agrawala, M. (2023).
\newblock Adding conditional control to text-to-image diffusion models.
\newblock {\em arXiv\/}.

\bibitem[Zhang {\em et~al.}(2020)Zhang, Zhang, Chen, Yuan, and
  Wen]{zhang2020cross}
Zhang, P., Zhang, B., Chen, D., Yuan, L., and Wen, F. (2020).
\newblock Cross-domain correspondence learning for exemplar-based image
  translation.
\newblock In {\em CVPR\/}, pages 5143--5153.

\bibitem[Zhang {\em et~al.}(2022)Zhang, Huang, Tang, Huang, Ma, Dong, and
  Xu]{zhang2022inversion}
Zhang, Y., Huang, N., Tang, F., Huang, H., Ma, C., Dong, W., and Xu, C. (2022).
\newblock Inversion-based creativity transfer with diffusion models.
\newblock {\em arXiv\/}.

\bibitem[Zhou {\em et~al.}(2021)Zhou, Zhang, Zhang, Zhang, Bao, Chen, Zhang,
  and Wen]{zhou2021cocosnet}
Zhou, X., Zhang, B., Zhang, T., Zhang, P., Bao, J., Chen, D., Zhang, Z., and
  Wen, F. (2021).
\newblock Cocosnet v2: Full-resolution correspondence learning for image
  translation.
\newblock In {\em CVPR\/}.

\end{thebibliography}
}

\newpage
\appendix

\end{document}